\documentclass[journal]{IEEEtran}
\usepackage{amsmath,amsfonts}
\usepackage{algorithmic}
\usepackage{array}
\usepackage[caption=false,font=normalsize,labelfont=sf,textfont=sf]{subfig}
\usepackage{textcomp}
\usepackage{stfloats}
\usepackage{url}
\usepackage{verbatim}
\usepackage{graphicx}
\usepackage{caption}
\usepackage[utf8]{inputenc} 
\usepackage[T1]{fontenc}
\usepackage{float} 
\usepackage{booktabs}
\usepackage{cite}
\usepackage{multirow}
\usepackage{diagbox}
\usepackage{makecell}
\usepackage[backref=none]{hyperref} 
\usepackage{subcaption}
\usepackage[font=small,labelfont=bf]{caption} 

\usepackage{balance}
\begin{document}

\title{Multi-Domain Features Guided Supervised Contrastive Learning for Radar Target Detection}

\author{Junjie Wang, Yuze Gao, Dongying Li, Wenxian Yu

\thanks{
(\textit{Corresponding author: Dongying Li.})

Junjie Wang, Yuze Gao, Dongying Li, Wenxian Yu are with the Shanghai Key Laboratory of Intelligent Sensing and Recognition,
Shanghai Jiao Tong University, Shanghai, 200240, China (email: iris\_1001@sjtu.edu.cn; gaoyuze@sjtu.edu.cn; dongying.li@sjtu.edu.cn; wxyu@sjtu.edu.cn).

} }

\maketitle

\begin{abstract}
Detecting small targets in sea clutter is challenging due to dynamic maritime conditions. Existing solutions either model sea clutter for detection or extract target features based on clutter-target echo differences, including statistical and deep features. While more common, the latter often excels in controlled scenarios but struggles with robust detection and generalization in diverse environments, limiting practical use.
In this letter, we propose a multi-domain features guided supervised contrastive learning method (MDFG\_SCL), which integrates statistical features derived from multi-domain differences with deep features obtained through supervised contrastive learning, thereby capturing both low-level domain-specific variations and high-level semantic information. This comprehensive feature integration enables the model to effectively distinguish between small targets and sea clutter, even under challenging conditions. Experiments conducted on real-world datasets demonstrate that the proposed shallow-to-deep detector not only achieves effective identification of small maritime targets but also maintains superior detection performance across varying sea conditions, outperforming the mainstream unsupervised contrastive learning and supervised contrastive learning methods.
\end{abstract}


\section{Introduction}

Radar target detection in sea clutter is essential for applications such as maritime surveillance and search and rescue. Recent advancements in deep learning improve robustness and generalization, providing an effective solution to the challenges posed by high power, non-homogeneity, and non-stationarity of sea clutter.

Researchers have initiated the exploration of target detection methods using machine learning models by identifying the distinctive characteristics of sea clutter and target echoes in the time domain, frequency domain, and time-frequency domain. Shui et al. \cite{shui2014tri} combined one time-domain feature and two frequency-domain features to propose a detector based on these three characteristics. Shi et al. \cite{shi2018sea} proposed a detector using the normalized Smoothed Pseudo Wigner-Ville Distribution (SPWVD) to extract time-frequency features from radar echoes.

Due to sea spikes, distinguishing between sea clutter and target echoes using certain characteristics is difficult. Therefore, target detection is often performed in high-dimensional feature space to improve performance. Wang et al. \cite{wang2022maritime} proposed a maritime radar detection method using CNN and Dual-Perspective Attention (DPA), encoding radar echoes in high-dimensional space and extracting global and local features. Chen et al. \cite{chen2024pointnet} introduced a PointNet-based approach for detecting multiple targets, integrating global and local features for both coarse and fine detection.

In addition to traditional supervised learning, semi-supervised, unsupervised, and contrastive learning approaches have advanced radar target detection. Jing et al. \cite{jing2022radar} introduced contrastive learning loss for improved detection on time-frequency plots. Xia et al. \cite{xia2023target} enhanced SimCLR with unlabeled radar echo signals for data augmentation, achieving higher detection accuracy with limited labeled data. Wang et al. \cite{wang2023maritime} proposed a self-evolving framework for semi-supervised radar detection, improving generalization capabilities.

To address the problem of decreased target detection performance caused by strong sea clutter, and poor generalization, inspired by \cite{xia2023target}, building on the basis of contrastive learning-based radar visual representation, we propose a radar target detection method guided by multi-domain shallow features for representation learning of high-dimensional features, which is named "MDFG-SCL". In this paper, we segment radar echo pulses into short segments and extract both multi-domain shallow and high-dimensional deep features. To address the varying distinctiveness of shallow features between sea clutter and target echoes, we use the Gini coefficient from CART to weight six features: Relative Average Amplitude (RAA), Relative Doppler Peak Height (RDPH), Relative Doppler Vector Entropy (RVE), Ridge Integral (RI), Number of Connected Regions (NR), and Maximum Size of Connected Regions (MS). For high-dimensional features, we hypothesize that features from different transformations should be similar, and features within the same category should align. A supervised contrastive loss function is used to implement this, along with a matching loss function to align shallow and deep features.

The main contributions of this paper are as follows:

\begin{enumerate}

\item A one-dimensional convolution-based supervised contrastive learning framework has been introduced for radar target detection, effectively distinguishing sea clutter and target echoes for improved performance.

\item Differential features between sea clutter and target echoes are utilized as multi-domain shallow features to guide the learning of high-dimensional deep features, significantly improving detection credibility and generalization.

\end{enumerate}

\section{Methodology}

\begin{figure}[htbp]
  \centering
  \includegraphics[width=1.0\linewidth]{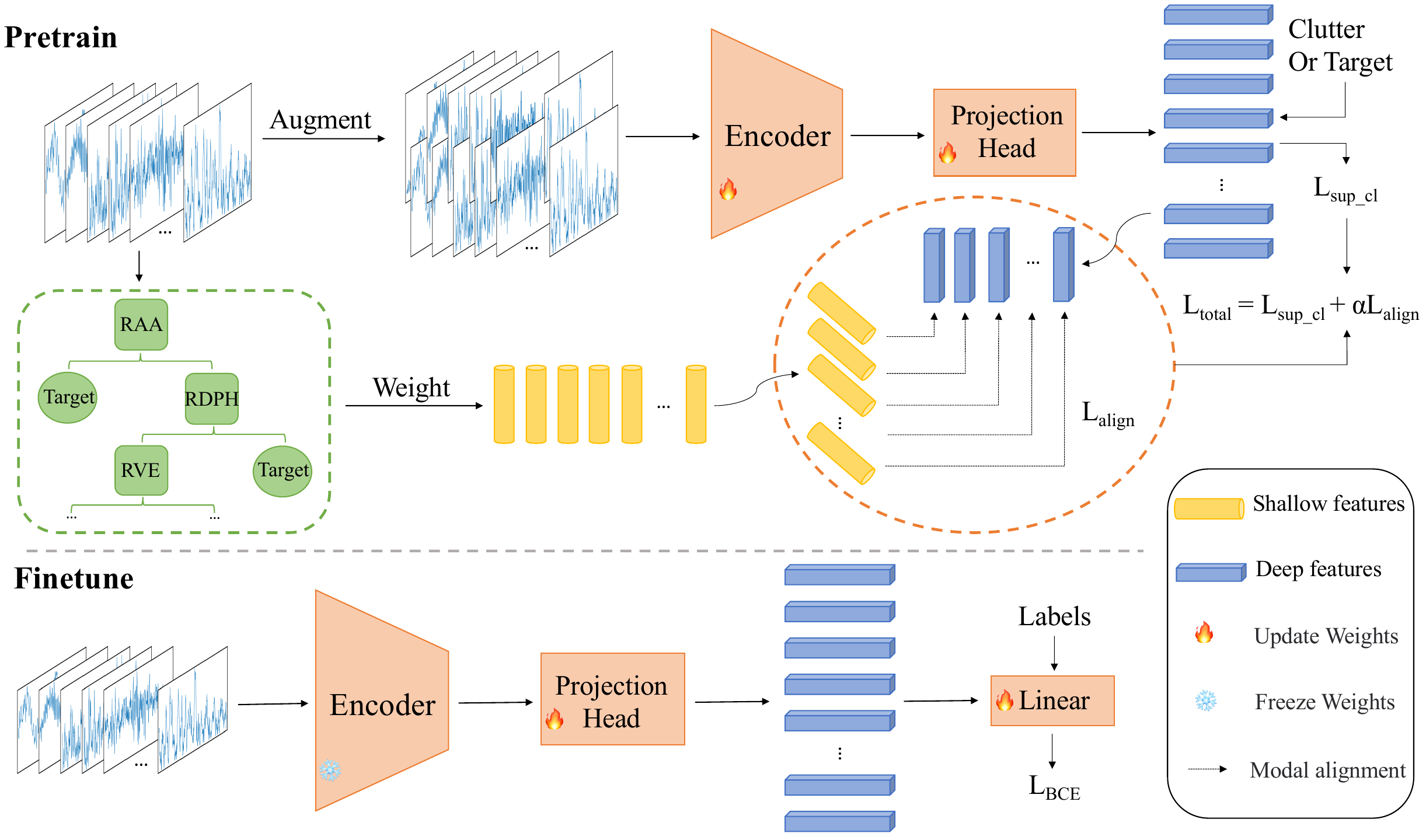} 
  \caption{ Supervised contrast learning framework based on multi-domain shallow features guidance.}
  \label{fig1}
\end{figure}

A supervised contrastive learning framework guided by shallow multi-domain features, as shown in Fig. \ref{fig1}., is proposed, consisting of pre-training and fine-tuning stages. Pre-training involves data augmentation, feature learning (deep and shallow), and loss functions like contrastive and matching losses. During fine-tuning, the pre-trained encoder is frozen, and only the projection head and classifier are trained.

\subsection{Multi-domain features extraction}

 Six commonly used distinctive features in sea surface target detection are selected, which include RAA in the time domain, RDPH and RVE in the frequency domain, and RI, NR, MS in the time-frequency domain. The six distinctive features mentioned above are used to calculate the multi-domain shallow features of the echo signal $ \boldsymbol{F}_{shallow}=[f_{RAA},f_{RDPH},f_{RVE},f_{RI},f_{NR},f_{MS}] $. The feature set $\boldsymbol{F}_{shallow}$ is normalized, and then the Gini index is calculated for each feature separately to assess the importance of each feature in distinguishing between sea clutter and target echoes.
 
The Gini index is used to measure the purity of a dataset \( D \) after classification with feature \( a \). First, the dataset \( D \) is divided into \( V \) subsets \( D_v \) based on feature \( a \), where each subset contains \( |D_v| \) samples, and \( p_{kv} \) represents the proportion of the \( k \)-th class in subset \( D_v \). The Gini index for feature \( a \) in dataset \( D \) is calculated as 

\begin{equation}
{Gini}(D, a) = \sum_{v=1}^{V} \frac{|D_v|}{|D|} \left( 1 - \sum_{k=1}^{K} p_{kv}^2 \right).
\end{equation}

The change in Gini index, \( \Delta \text{Gini}(D, a) = \text{Gini}(D) - \text{Gini}(D, a) \), is used to assess the contribution of feature \( a \) to the classification task. A larger \( \Delta \text{Gini} \) indicates that feature \( a \) has a greater impact on the classification, implying higher importance.

By calculating the Gini index for each feature in the dataset and ranking the six multi-domain shallow features by importance, the features are weighted accordingly. This ensures that features with higher weights have a stronger influence on the deep feature representation.

\subsection{Supervised contrastive learning method for radar target detection}

\subsubsection{Data Augmetation}

Data augmentation is performed on the input complex signal using methods outlined in \cite{xia2023target}, which have been previously validated for their effectiveness on radar echo signals. The methods include the Random Cropping and Resampling Data Augmentation Method (RCRS-DAM), the Adding Disturbance Data Augmentation Method (AD-DAM), and the Flipping Data Augmentation Method (F-DAM).

Each complex-valued sample is augmented using two of the three methods mentioned, along with its original form. The modulus of the transformed complex signal is taken, converting it to real-valued data, which is then input into the encoder for deep feature extraction.

\subsubsection{Deep Feature Extraction}

After augmentation, the echo segments are processed by the encoder and projection head to extract their deep features. In this paper, we utilize a 1D-Resnet50 \cite{he2016deep} as the encoder to capture the representation of the signals. A two-layer MLP is used to map the high-dimensional representation output from the encoder to a relatively lower-dimensional space, defined as:

\begin{equation}
\boldsymbol{z}=g(\boldsymbol{x})=\boldsymbol{W}_{2}\mathrm{ReLU}(\boldsymbol{W}_{1}\boldsymbol{x}+\boldsymbol{b1})+\boldsymbol{b2}
\end{equation}

where $\boldsymbol{W_1}$ and $\boldsymbol{W_2}$ represent the learnable weights for the linear layers, $\boldsymbol{b_1}$ and $\boldsymbol{b_2}$ represent the bias for the linear layers, and ReLU(·) denotes the rectified linear unit (ReLU) function.

\subsubsection{Match Loss Function and Supervised Contrastive Loss Function}

A matching loss function has been proposed to align shallow features with deep features modally. Shallow and deep features are mapped to the same feature space, where a similarity matrix is computed to determine the distances between corresponding shallow and deep features. Subsequently, cross-entropy loss is calculated for both the shallow feature dimension and the deep feature dimension to guide the high-dimensional deep features towards the shallow features. Finally, an average is taken to obtain the matching loss function. The respective calculations are defined as

\begin{equation} \boldsymbol{d} = \frac{\boldsymbol{W}_d^T\boldsymbol{f}_d}{||\boldsymbol{W}_d^T\boldsymbol{f}_d||} \boldsymbol{s} = \frac{\boldsymbol{W}_s^T\boldsymbol{f}_s}{||\boldsymbol{W}_s^T\boldsymbol{f}_s||} \label{equation2}\end{equation}

where $\boldsymbol{f}_d$ represents the deep features extracted via the encoder and projection head, while $\boldsymbol{f}_s$ are the weighted shallow features. $\boldsymbol{W}_d$ and $\boldsymbol{W}_s$ respectively are the transformation matrices for deep and shallow features. Through the mapping transformation given by the  Equation (\ref{equation2}), both $\boldsymbol{d}$ and $\boldsymbol{s}$ are transformed into $m$-dimensional normalized features within an implicit space. The matching loss function is defined as

\begin{equation}
    \begin{split}
    L_{align} = -\frac{1}{2N}( \sum_{i}^N log \frac{exp(\boldsymbol{s}_i^T \boldsymbol{d}_i)}{\sum_{j=1}^N exp(\boldsymbol{s}_i^T \boldsymbol{d}_j)}  + \\ \sum_{i}^N log \frac{exp(\boldsymbol{d}_i^T \boldsymbol{s}_i)}{\sum_{j=1}^N exp(\boldsymbol{d}_i^T \boldsymbol{s}_j)}) 
    \end{split}\label{equation3}
\end{equation}

where \( N \) represents the batch size. 

The supervised contrastive loss function, in the presence of labels, involves selecting any positive sample as the anchor and calculating the distance between it and all other positive samples, defined as

\begin{equation} L_{sup\_cl} = -\sum_{i}^N \frac{1}{P} \sum_{p}^P log \frac{exp(\boldsymbol{z}_i  \boldsymbol{z}_p/t)}{\sum_{a=1}^A exp(\boldsymbol{z}_i  \boldsymbol{z}_a/t)}  p \neq i,a \neq i,\end{equation}

where $N$ is the total number of samples, $P$ is the number of positive samples that belong to the same category as sample $\boldsymbol{z}_i$, $ \boldsymbol{z}_p$ is a positive sample of $ \boldsymbol{z}_i$, $A$ is the number of negative samples that belong to different classes than sample $\boldsymbol{z}_i$, and $\boldsymbol{z}_a$ is a negative sample of $\boldsymbol{z}_i$, $t$ is temperature coefficient.

During the model pre-training process, both the match loss and the supervised contrastive loss simultaneously optimize the parameters of the encoder and the projection head. Therefore, the total training loss of the model is defined as 

\begin{equation} 
L_{total} = L_{sup\_cl} + \alpha L_{align},
\end{equation}

where $\alpha$ is a coefficient used to balance the modal alignment loss. 


\section{Experiments and Discussion}

\subsection{Datasets Description}

This letter uses the internationally recognized Intelligent PIXel (IPIX) radar sea clutter dataset \cite{IPIX} for pre-training and testing, collected by McMaster University in 1993. The dataset details are shown in Table 1. The dataset consists of 14 range cells, categorized into primary target cells, secondary target cells, and pure clutter cells, with this study focusing on primary target cells. Echo signals are received through both horizontal and vertical polarization channels, resulting in four polarization states: HH, HV, VH, and VV.

\begin{table}[htbp]
  \centering
  \caption{Description of ten datasets in IPIX1993}
  \resizebox{8.5cm}{!}{
    \begin{tabular}{l|ccccc}
    \toprule
    Data id & Primary target cells & Secondary target cells & SCR of HH/HV/VH/VV(dB)\\
    \midrule
    \#17  & 9     & 8,10,11  & 16.9/12.5/12.5/3.5\\
    \#26  & 7     & 6,8   & 4.3/5.9/5.9/5.7\\
    \#30  & 7     & 6,8   & -0.3/3.6/3.6/2.0\\
    \#31  & 7     & 6,8,9  & 6.5/7.4/7.4/8.2\\
    \#40  & 7     & 5,6,8  & 9.5/12.9/12.8/11.0\\
    \#54  & 8     & 7,9,10  & 18.0/16.1/16.2/8.8\\
    \#280 & 8     & 7,9,10  & 4.0/7.3/7.4/4.4\\
    \#310 & 7     & 6,8,9  & 2.3/5.0/5.0/-1.5\\
    \#311 & 7     & 6,8,9  & 11.9/14.7/14.7/8.7\\
    \#320 & 7     & 6,8,9  & 11.8/13.7/13.7/6.8\\
    \bottomrule
    \end{tabular}}
  \label{tab:1}
\end{table}

\subsection{Experiment Setting}

The IPIX radar dataset is used to assess the MDFG\_SCL, with each dataset divided into \( N = 512 \) segments for both target and clutter cells. Due to the limited target cells, oversampling results in approximately 2,000 target samples and 2,400 clutter samples per dataset. 80\% of the samples are used for pre-training in supervised contrastive learning, and the remainder is split into training, validation, and testing sets in a 2:2:1 ratio. Only pure clutter samples are used for the validation set to calculate the detection threshold based on a given false alarm rate $P_{fa}$. Experiments are conducted on a server with an NVIDIA GeForce RTX 4070 Super GPU and the PyTorch framework. During pre-training, a batch size of 128 and 100 epochs are used, with an SGD optimizer, a learning rate of 0.01, and weight decay of 1e-4. The contrastive learning temperature coefficient is set to 0.07.

\subsection{Results and Discussion}

The experiments are divided into multiple parts. First, the proposed MDFG\_SCL method is compared with state-of-the-art approaches on the IPIX1993 dataset. Then, ablation studies are conducted on key modules of MDFG\_SCL, including supervised contrastive learning, shallow feature modules, and the matching loss coefficient $\alpha$. Finally, detection results are visualized to compare the generalization performance of different methods using different datasets for pre-training and testing.

To evaluate detection experiments quantitatively, five metrics—accuracy, precision, recall ($P_{d}$), false alarm rate, and mIoU—are calculated from the classification confusion matrix, as defined in Table \ref{tab:2}.

\begin{table*}[htbp]
  \centering
  \caption{Confusion Matrix and Evaluation Metrics}
    \begin{tabular}{c|cc||ccccc}
    \toprule
    \multirow{2}[0]{*}{\diagbox{Actual}{Predicted}} & \multirow{2}[0]{*}{Clutter} & \multirow{2}[0]{*}{Target} & \multirow{2}[0]{*}{Accuracy} & \multirow{2}[0]{*}{Precision} & \multirow{2}[0]{*}{Recall} & \multirow{2}[0]{*}{True $P_{fa}$} & \multirow{2}[0]{*}{mIOU} \\
          &       &       &       &       &       &       &  \\
    \midrule
    Clutter & TN    & FP    & \multirow{2}[0]{*}{$\frac{\mathrm{TP+TN}}{\mathrm{TP+FN+FP+TN}}$} & \multirow{2}[0]{*}{$\frac{\mathrm{TP}}{\mathrm{TP}+\mathrm{FP}}$} & \multirow{2}[0]{*}{$\frac{\mathrm{TP}}{\mathrm{TP}+\mathrm{FN}}$} & \multirow{2}[0]{*}{$\frac{\mathrm{FP}}{\mathrm{FP}+\mathrm{TN}}$} & \multirow{2}[0]{*}{$\frac{1}{2}{\left(\frac{\mathrm{TP}}{\mathrm{TP}+\mathrm{FN}+\mathrm{FP}}+\frac{\mathrm{TN}}{\mathrm{TN}+\mathrm{FN}+\mathrm{FP}}\right)}$} \\
    Target & FN    & TP    &       &       &       &       &  \\
    \bottomrule
    \end{tabular}%
  \label{tab:2}%
\end{table*}%

\subsubsection{Performance Comprison}

\begin{figure}[htbp]
    \centering
    \subfloat[HH]{%
        \includegraphics[width=0.48\linewidth]{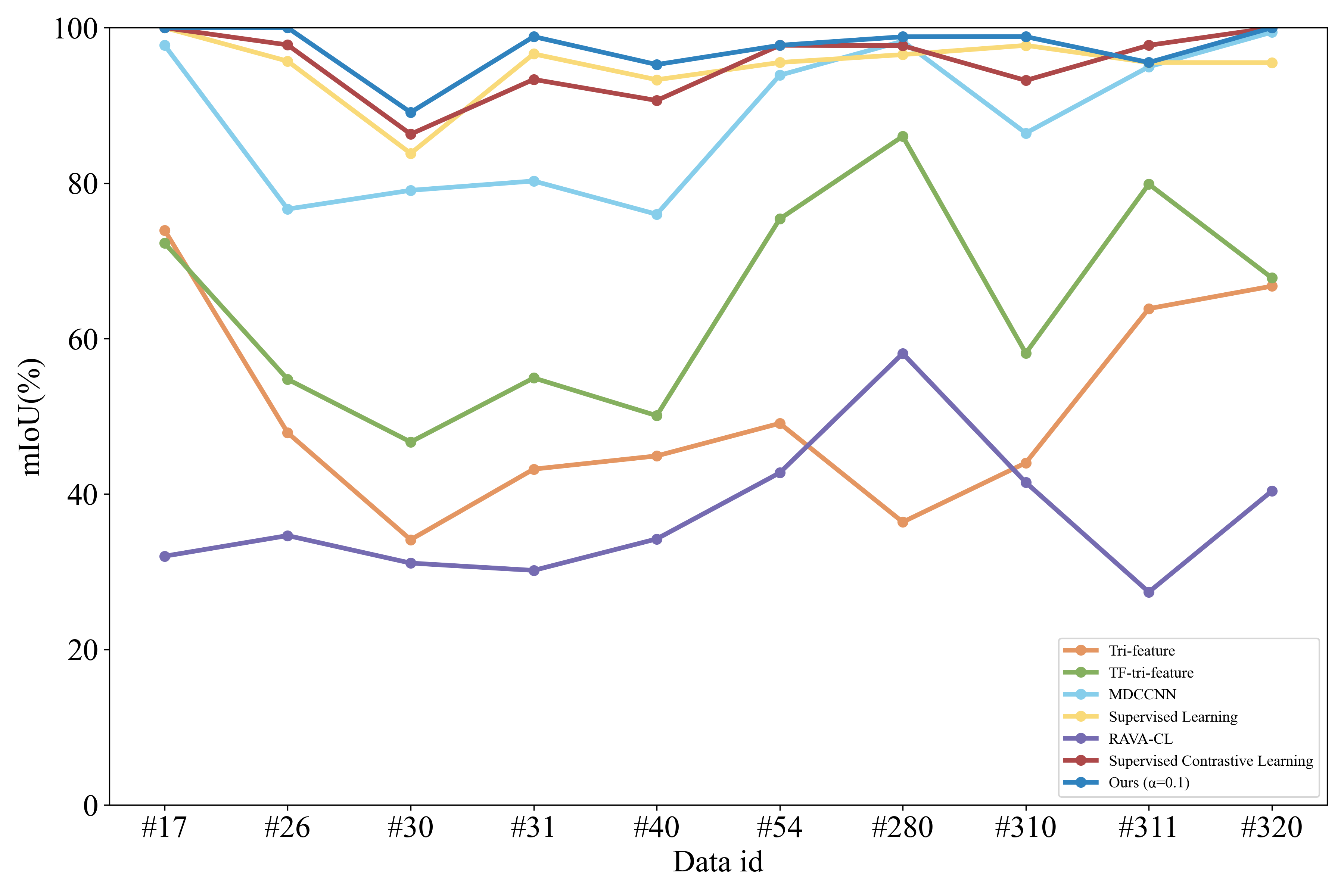}
    }
    \hfill
    \subfloat[HV]{%
        \includegraphics[width=0.48\linewidth]{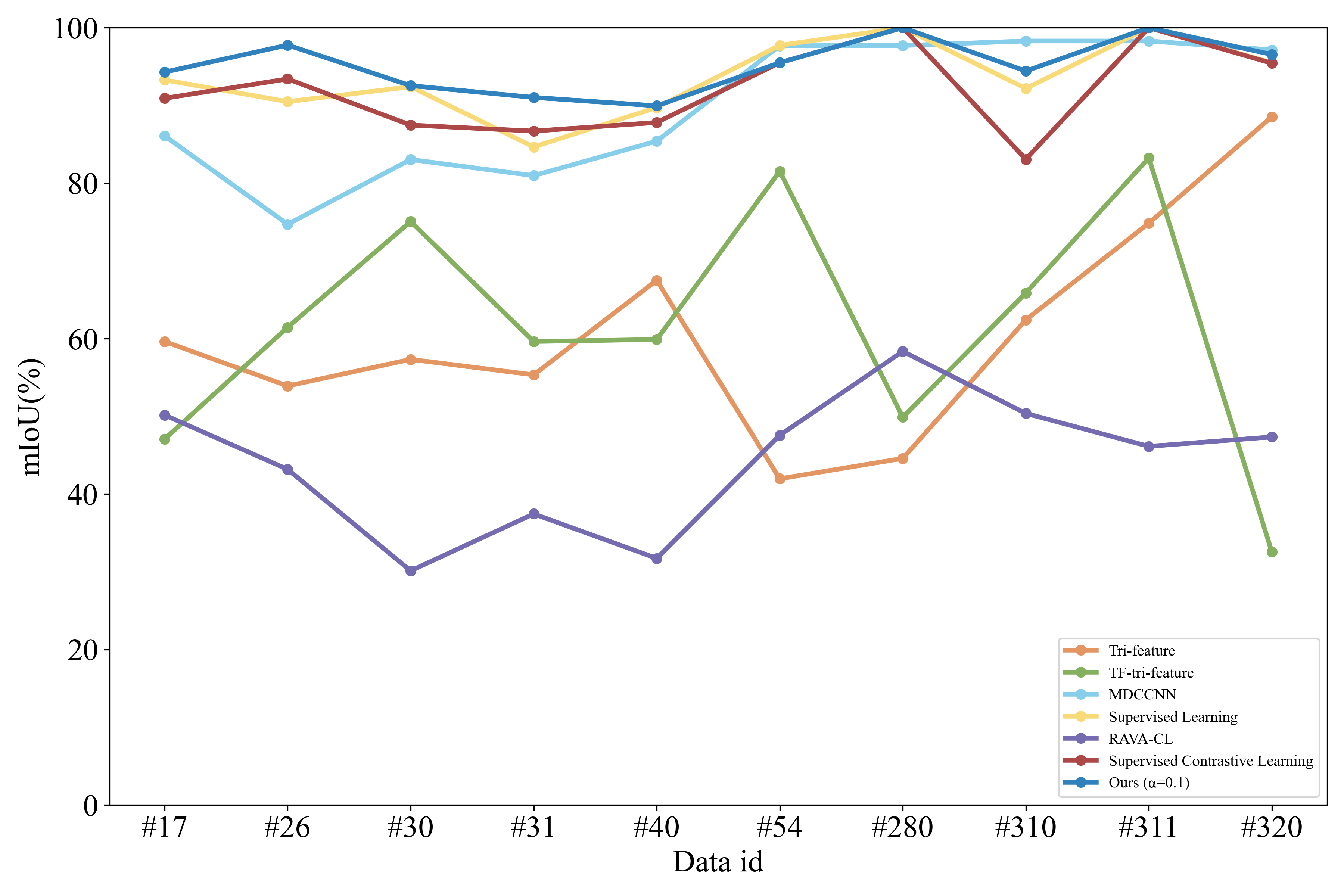}
       }\\[0pt]
    \subfloat[VH]{%
        \includegraphics[width=0.48\linewidth]{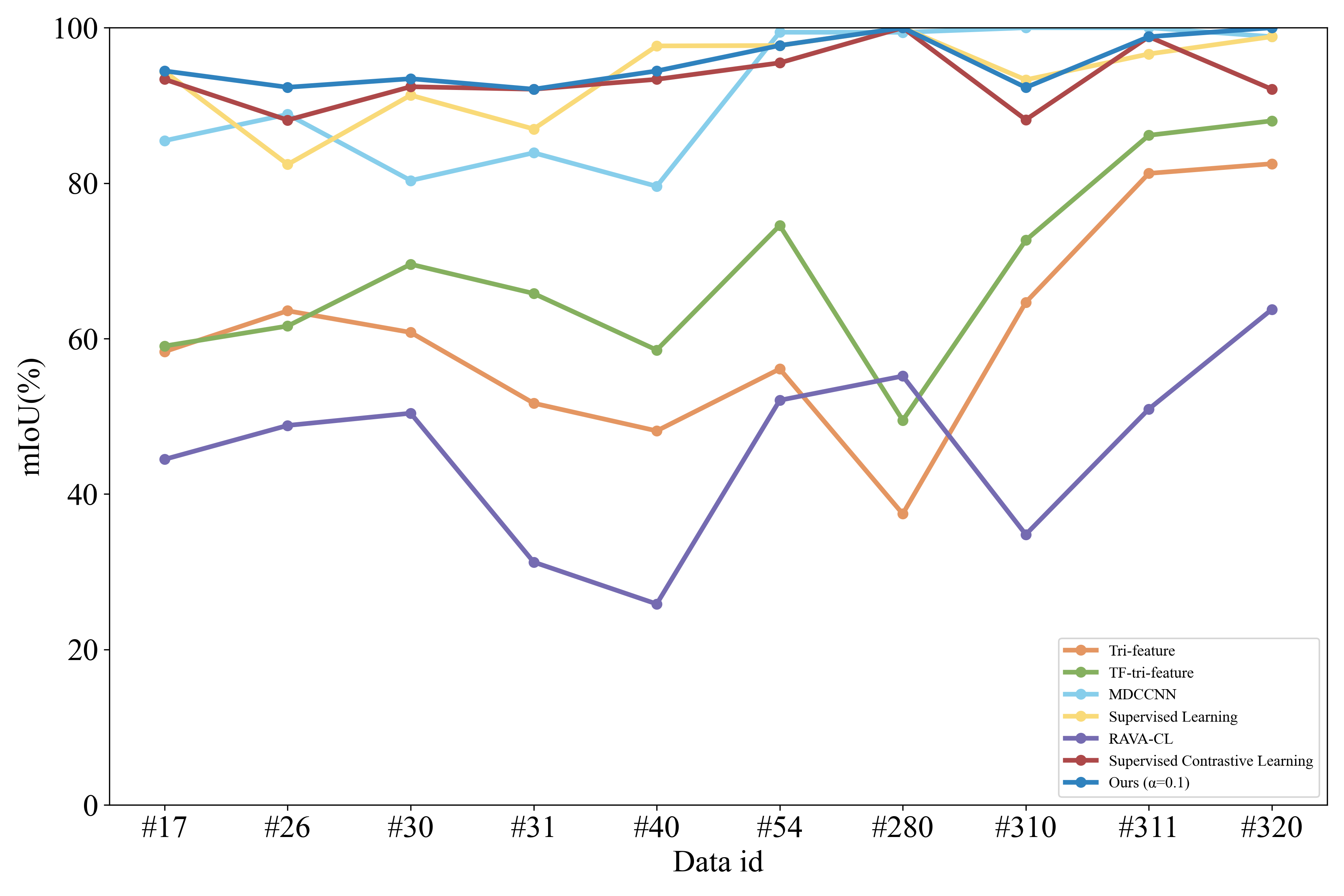}
    }
    \hfill
    \subfloat[VV]{%
        \includegraphics[width=0.48\linewidth]{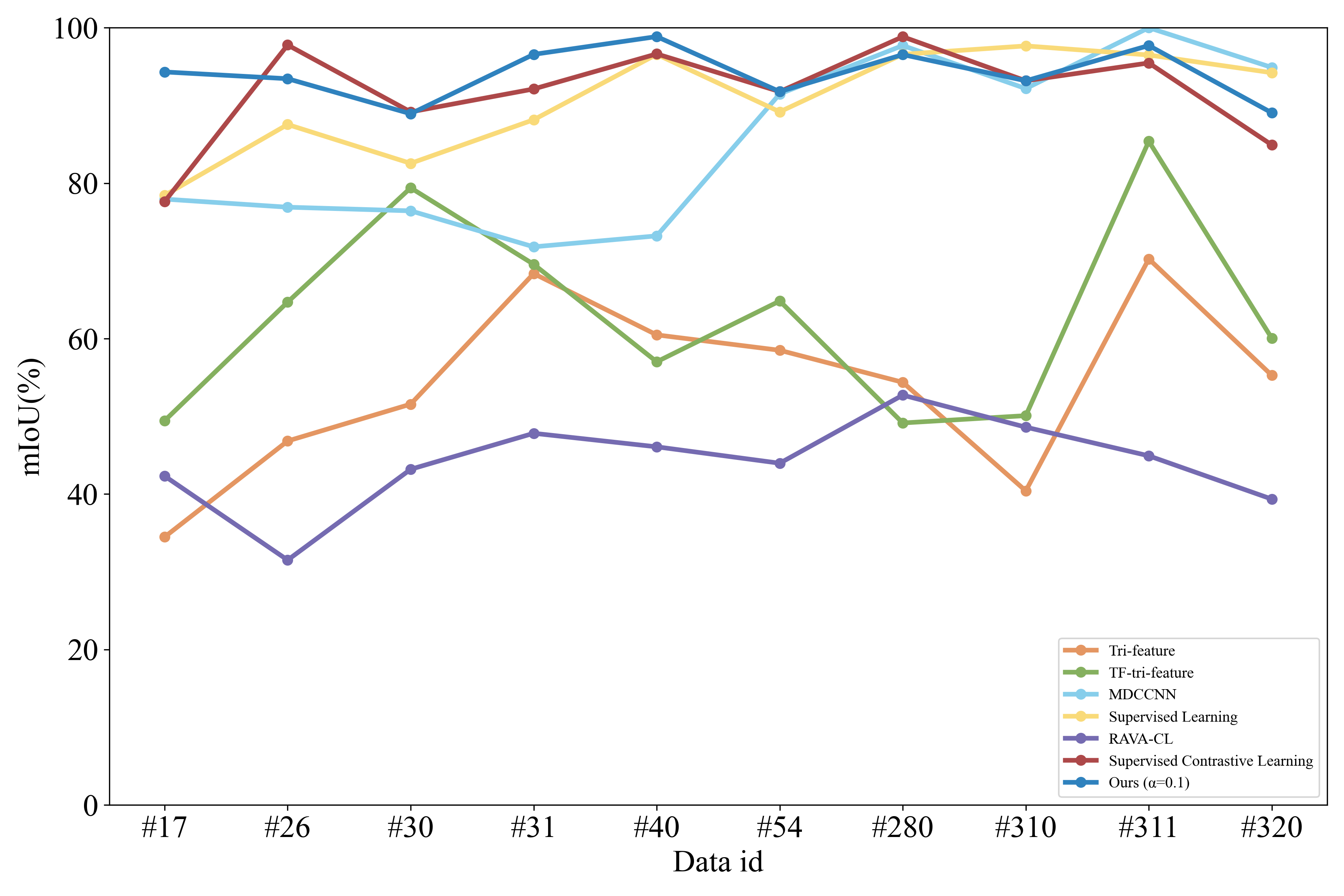}
    }
    
    \caption{Detection performance mIoU at $P_{fa}$=0.001.}
    \label{fig2}
\end{figure}

This section compares the proposed MDFG\_SCL method with six approaches: the Tri-feature detector \cite{shui2014tri}, TF-tri-feature detector \cite{shi2018sea}, MDCCNN detector \cite{chen2021false}, traditional supervised, unsupervised contrastive (RAVA-CL) \cite{xia2023target}, and supervised contrastive \cite{khosla2020supervised}. All methods use a 0.512s observation time. Detection performance mIoU at $P_{fa}$=0.001 is shown in Fig. \ref{fig2}, and the average true $P_{fa}$ for the ten datasets is listed in Table \ref{tab:3}.

\begin{table*}[htbp]
  \centering
  \caption{Average True $P_{fa}$ of Different Methods}
  \resizebox{\textwidth}{!}{
\begin{tabular}{l|ccccccc}
    \toprule
    Method & Tri-Feature\cite{shui2014tri} & TF-Tri-Feature\cite{shi2018sea} & MDCCNN\cite{chen2021false} & Supervised Learning & RAVA-CL\cite{xia2023target} & Supervised Contrastive Learning\cite{khosla2020supervised} & Ours ($\alpha=0.1$) \\
    \midrule
    True $P_{fa}$ & 0.1155 & 0.1676 & 0.0073 & 0.0047 & 0.0765 & 0.0033 & \textbf{0.0010} \\
    \bottomrule
    \end{tabular}%
  }
    
  \label{tab:3}%
\end{table*}%

As shown in Fig. \ref{fig2}, the proposed MDFG\_SCL method demonstrates robust detection performance across all four polarization modes and ten datasets, outperforming other detectors. In contrast, the Tri-feature detector \cite{shui2014tri} and TF-tri-feature detector \cite{shi2018sea}, which rely solely on shallow features, exhibit lower and unstable detection probabilities. The MDCCNN detector \cite{chen2021false} performs well on datasets with high signal-to-clutter ratios but suffers significant performance drops on low signal-to-clutter datasets. Traditional supervised methods and the supervised contrastive method \cite{khosla2020supervised} maintain relatively high detection performance due to the feature extraction capabilities of deep models. However, the unsupervised contrastive method RAVA-CL \cite{xia2023target} struggles due to the lack of supervised information during training, resulting in poor detection performance and difficulty achieving low false alarm rates. By integrating shallow and deep features, the proposed MDFG\_SCL consistently achieves strong detection performance under various sea conditions. As shown in Table \ref{tab:3}, MDFG\_SCL achieves the lowest false alarm rates while maintaining superior detection performance compared to other methods.

\subsubsection{Ablation Experiments}


\begin{table}[htbp]
  \centering
  \tabcolsep=0.08cm
  \caption{Ablation Experiment of Module}
    \begin{tabular}{l|ccccc}
    \toprule
    Method (Backbone=1D-Resnet50) & Accuracy(\%) & Recall(\%)  & mIOU(\%) \\
    \midrule
    Supervised Learning & 97.43 & 94.88 &  95.03 \\
    Supervised Contrastive Learning \cite{khosla2020supervised} & 97.70 & 95.59 &  95.45 \\
    Ours (without weights) & 98.21 & 96.54 &  96.40 \\
    Ours ($\alpha=0.1$) & \textbf{98.72} & \textbf{97.14} & \textbf{97.42} \\
    \bottomrule
    \end{tabular}%
  \label{tab:4}%
\end{table}%

To evaluate the effectiveness of each module in the proposed MDFG\_SCL for radar target detection, experiments using 1D-ResNet50 were conducted with different methods: traditional supervised learning, supervised contrastive learning, contrastive learning with multi-domain features, and with weighted multi-domain features. Results, averaged across ten IPIX1993 datasets in HH polarization, are shown in Table \ref{tab:4}.
The table shows that under the preset $P_{fa}$=0.001, only the proposed method meets the specified $P_{fa}$. Results demonstrate that the proposed MDFG\_SCL achieves a slight improvement with supervised contrastive learning compared to traditional supervised learning. Moreover, incorporating weighted multi-domain shallow features yields a 1\% increase in accuracy while achieving a lower false alarm rate. This aligns with the radar target detection requirement for high detection probability and stable low $P_{fa}$.

\begin{figure}[htbp]
      \centering
      \includegraphics[width=7.5cm]{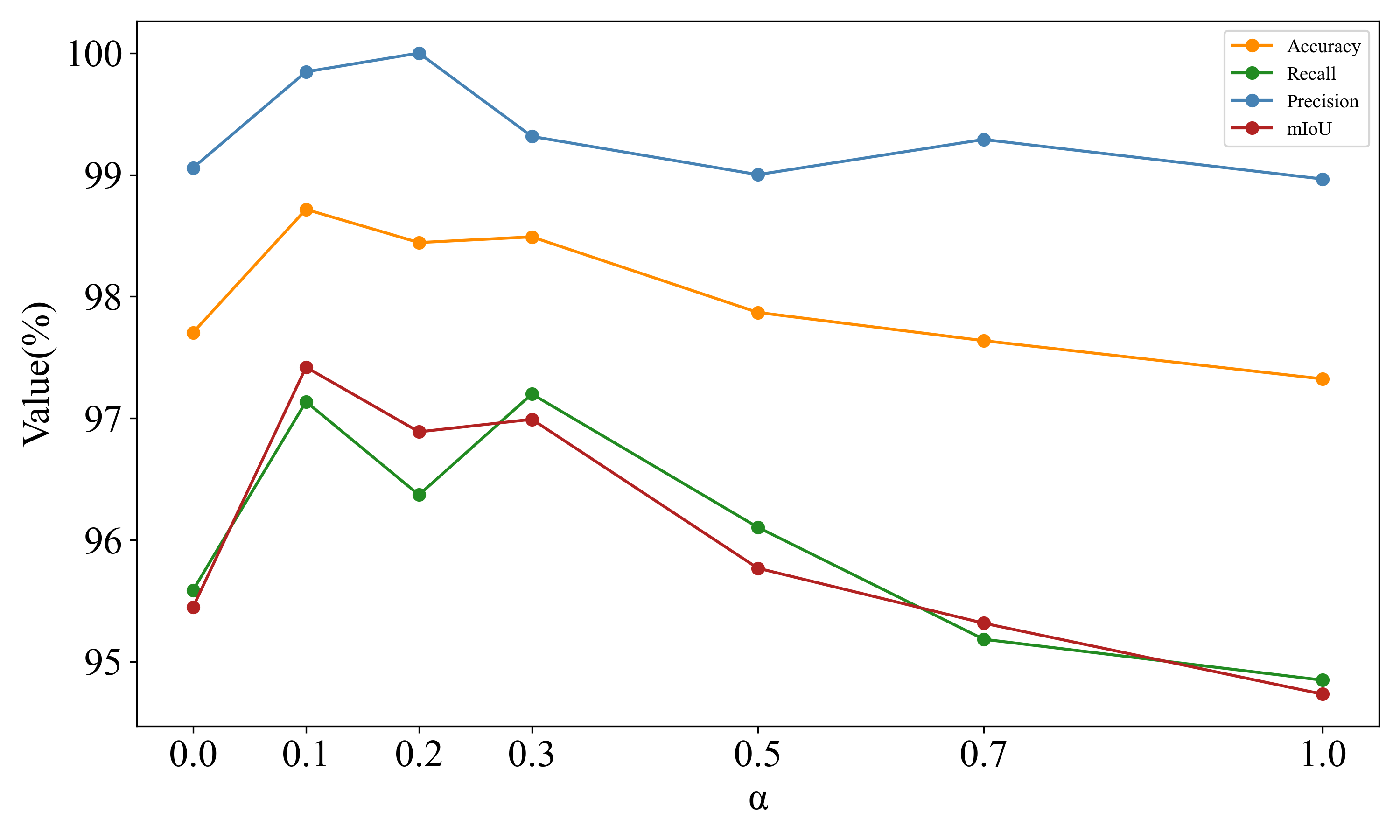} 
      \caption{Evaluation metrics change under different $\alpha$.}
      \label{fig3}
    \end{figure}

Experiments on ten datasets were conducted to determine the optimal \( \alpha \), which balances matching loss in the overall loss, with \( \alpha \) values of 0.1, 0.2, 0.3, 0.4, 0.7, 1, and \( \alpha \) = 0 as a baseline (supervised contrastive learning without shallow features). Average results are shown in Fig. \ref{fig3}.
Results show that when \( \alpha \) is between 0.1 and 0.3, detection performance improves compared to \( \alpha \) = 0, but declines when \( \alpha \) > 0.5, even falling below \( \alpha = 0 \) . Moderate shallow feature guidance enhances deep feature learning, while excessive \( \alpha \) overemphasizes shallow features, weakening the model. This explains the suboptimal performance of shallow features alone in prior studies. Other experiments use \( \alpha \) = 0.1.

\begin{figure*}[htbp]
    \centering
    \subfloat[]{%
        \includegraphics[width=0.24\linewidth]{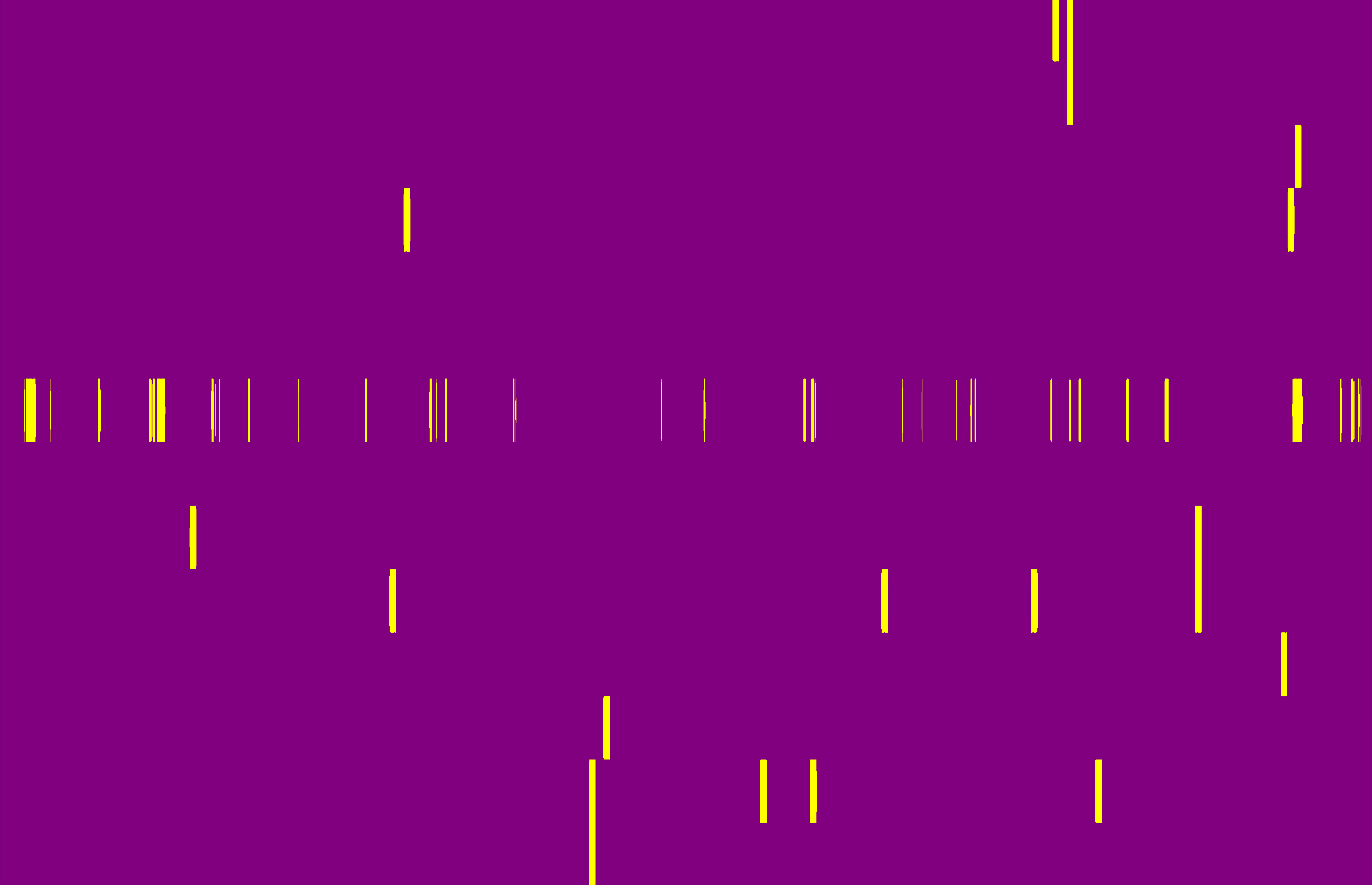}
        }
    \hfill
    \subfloat[]{%
        \includegraphics[width=0.24\linewidth]{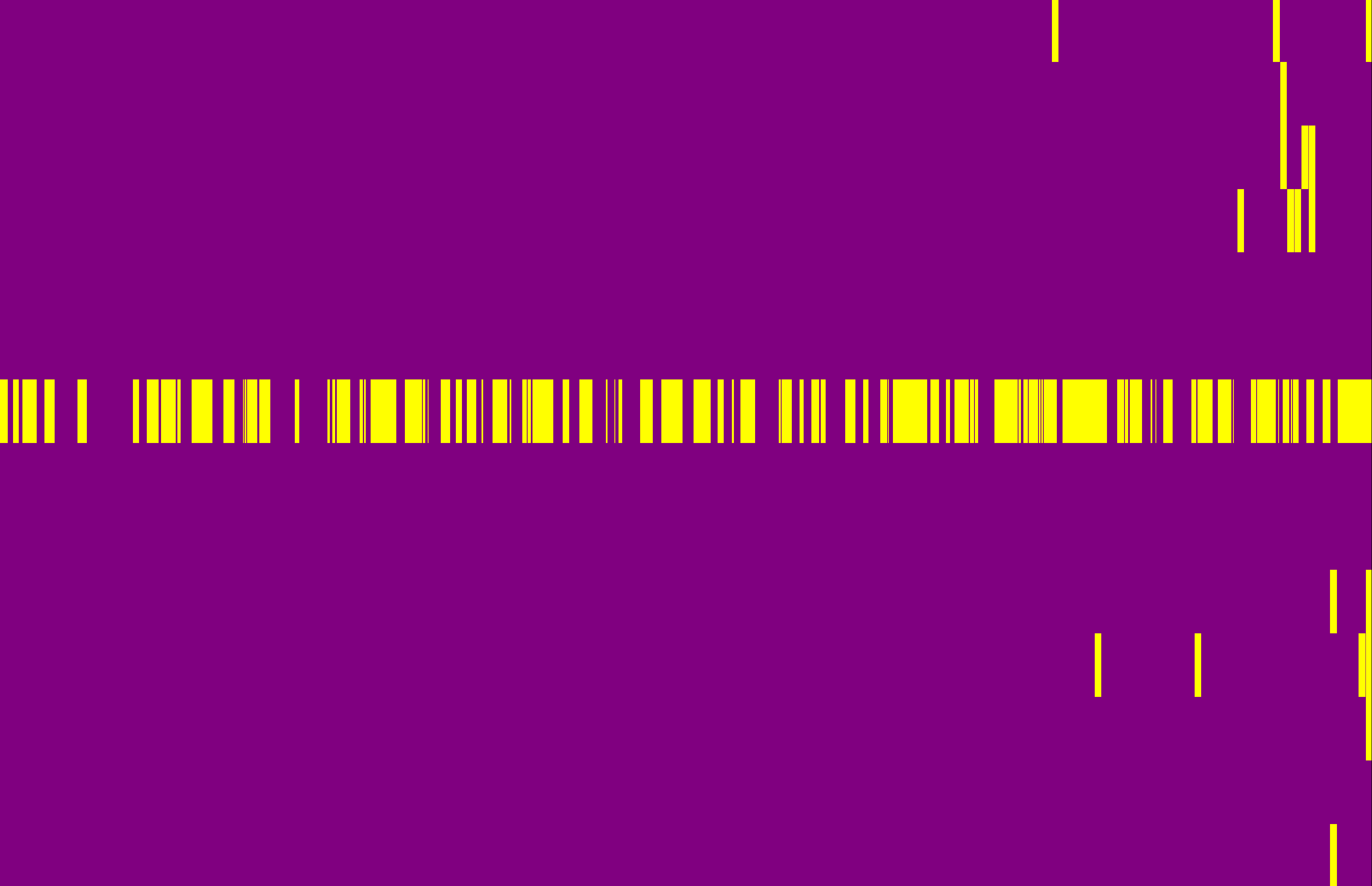}
        }
    \hfill
    \subfloat[]{%
        \includegraphics[width=0.24\linewidth]{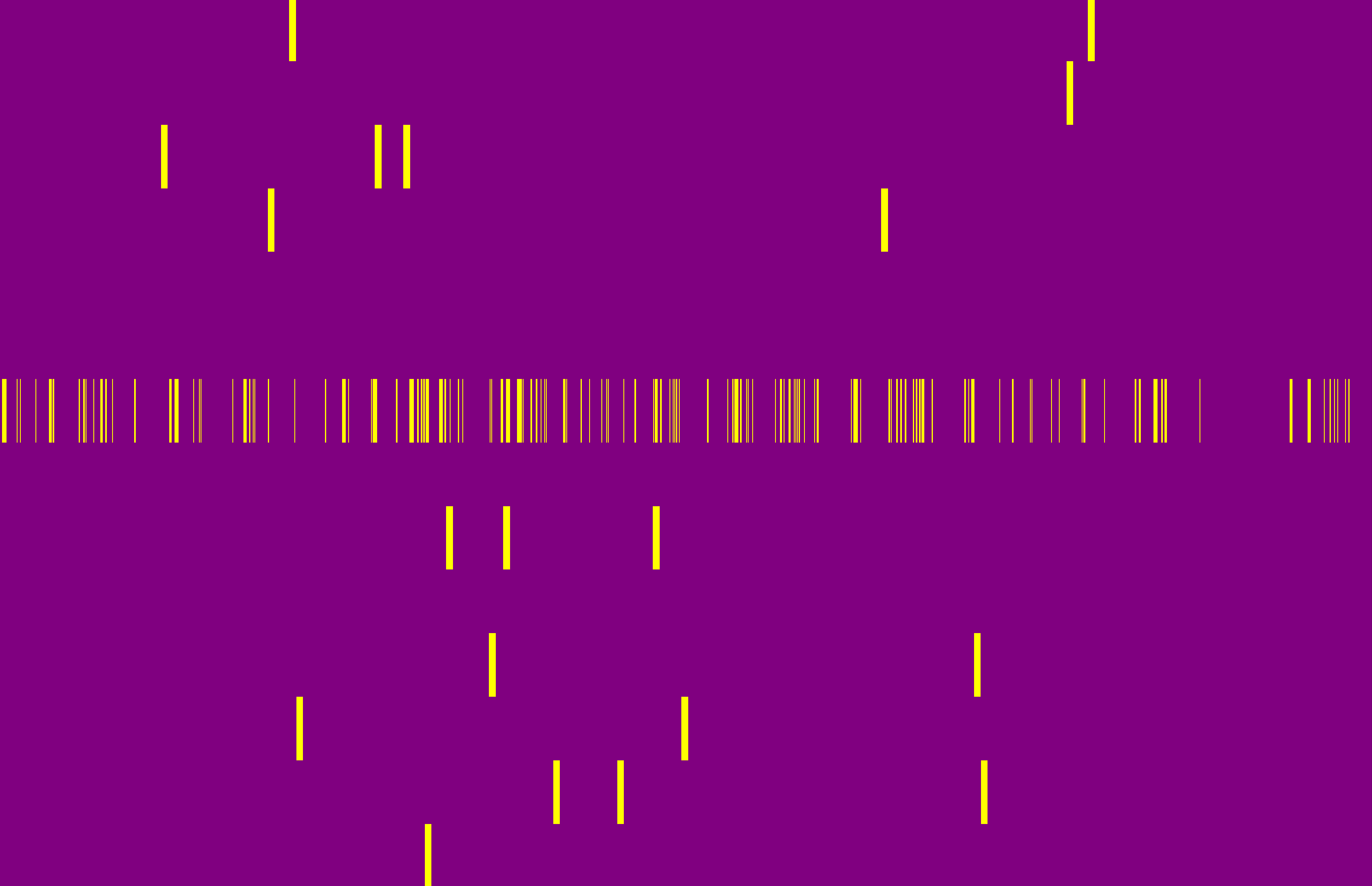}
       }
    \hfill
    \subfloat[]{%
        \includegraphics[width=0.24\linewidth]{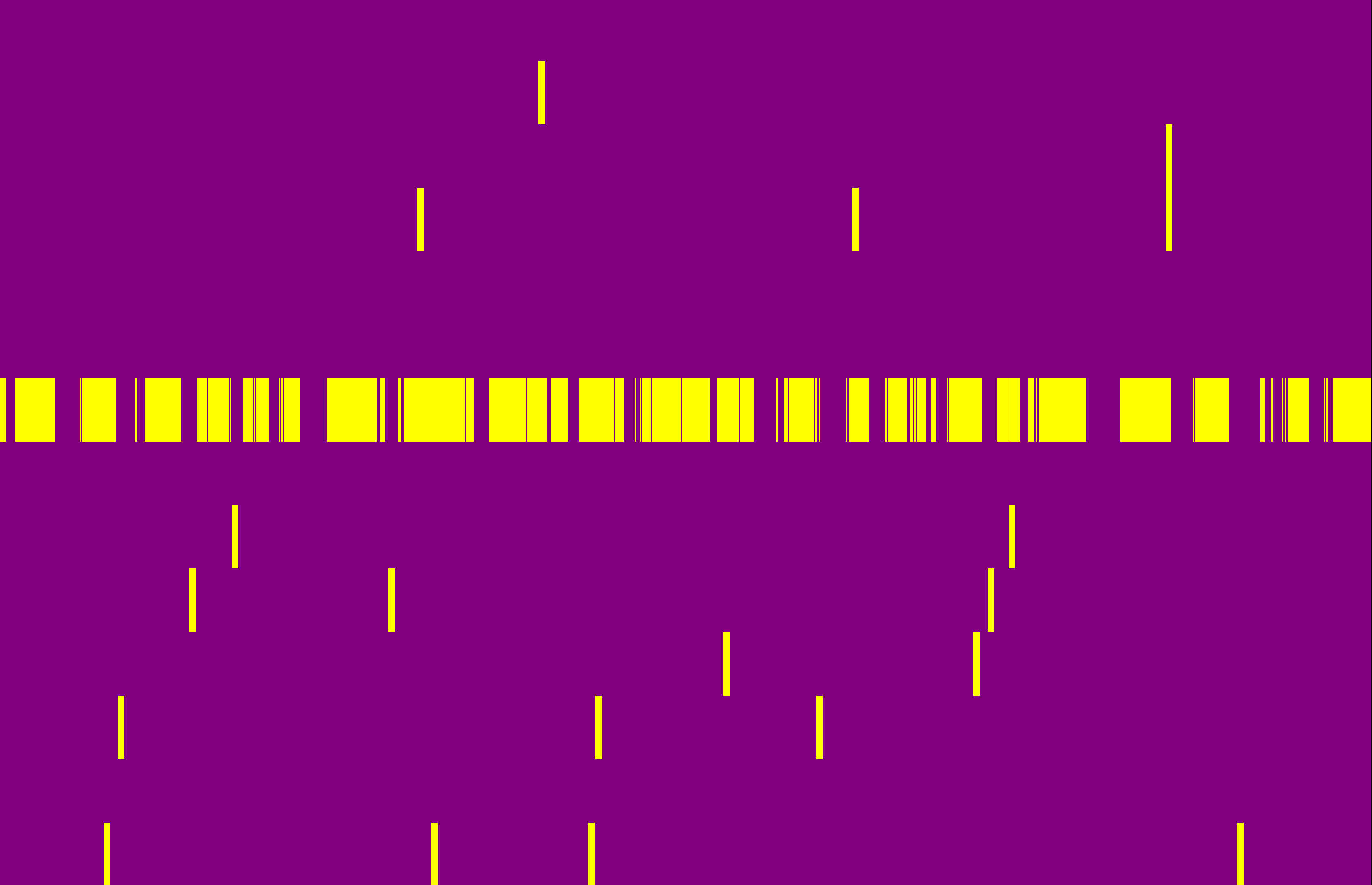}
       }\\[0pt]
    
    \subfloat[]{%
        \includegraphics[width=0.24\linewidth]{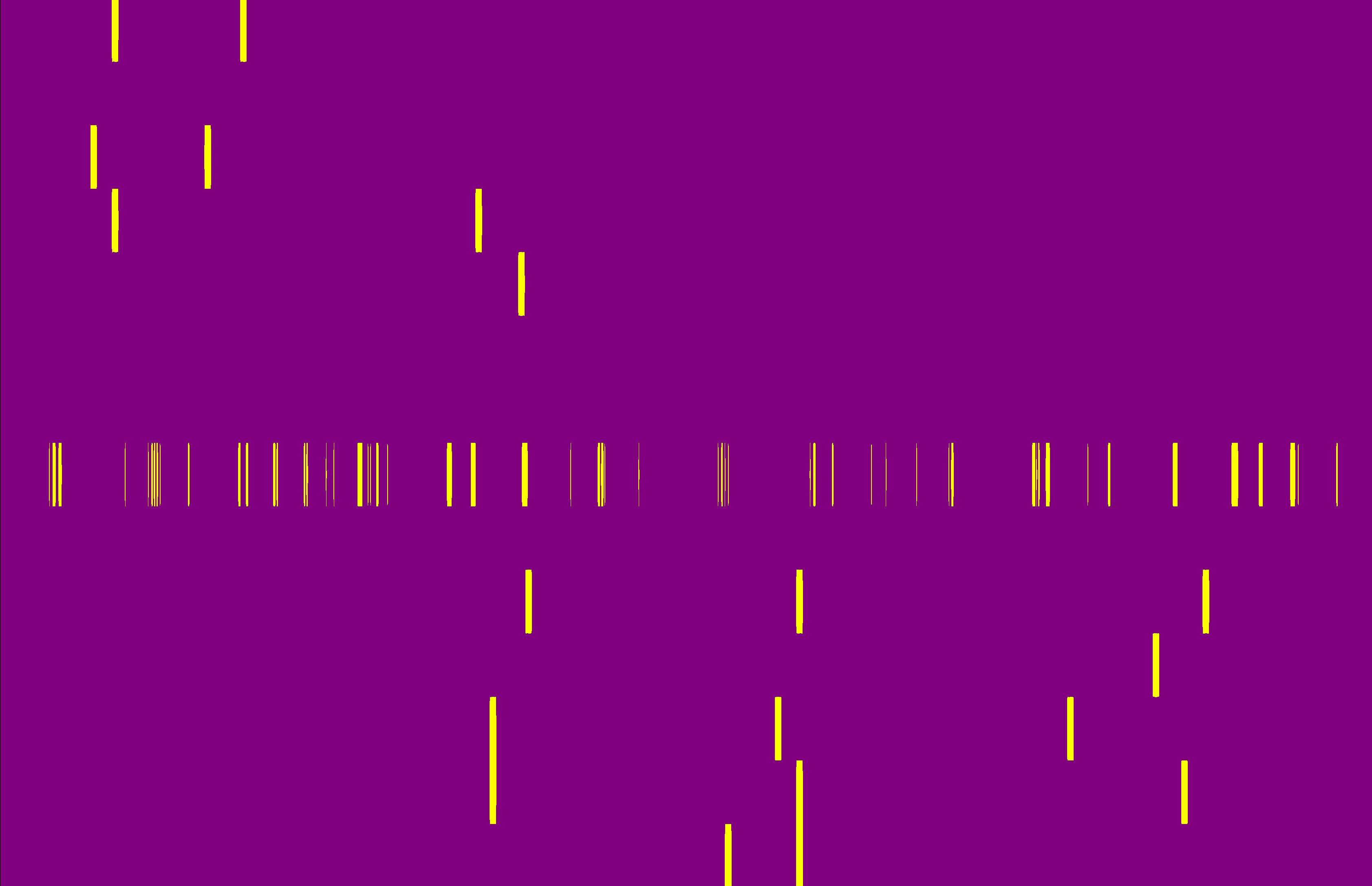}
       }
    \hfill
    \subfloat[]{%
        \includegraphics[width=0.24\linewidth]{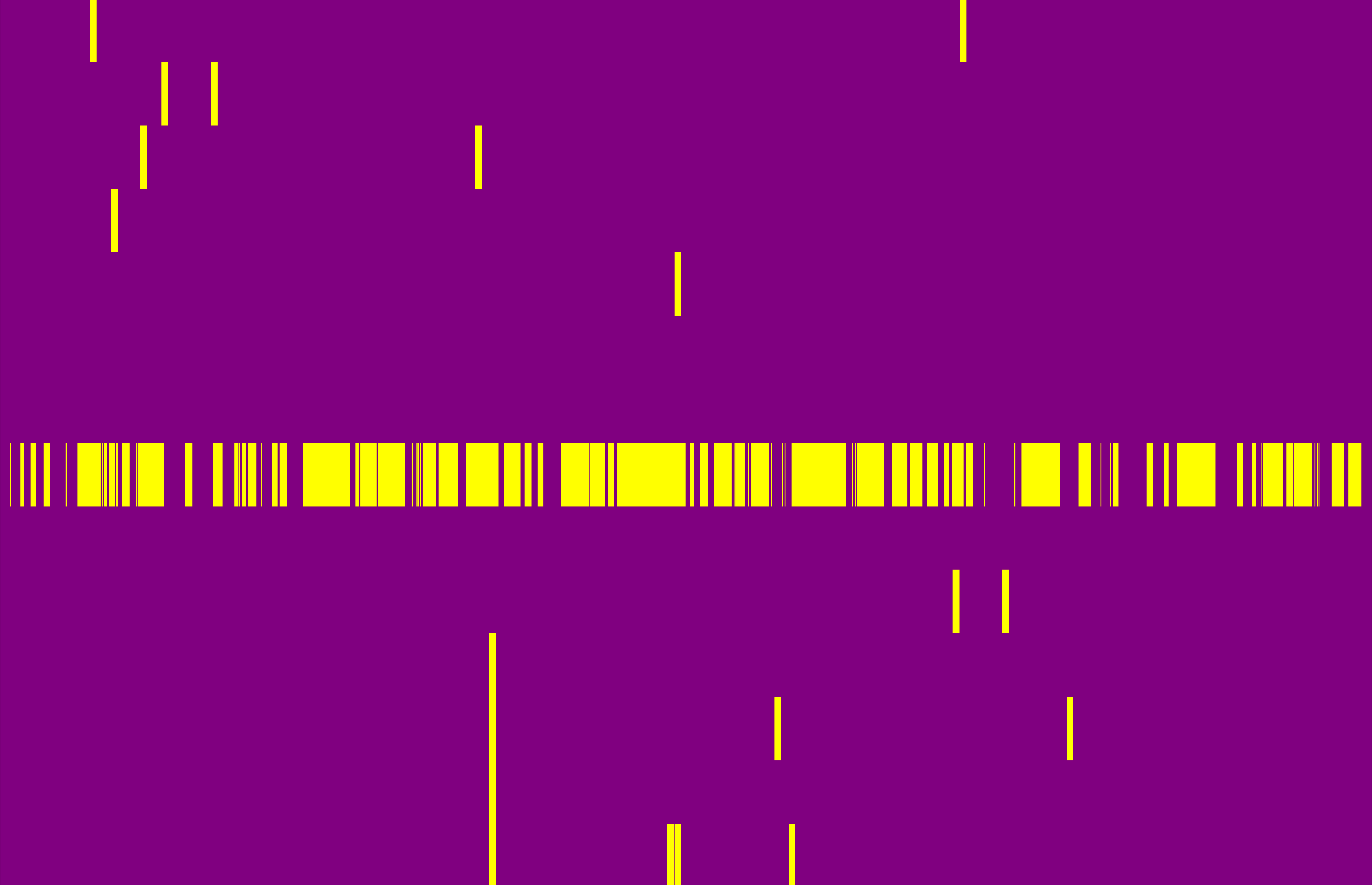}
       }
    \hfill
    \subfloat[]{%
        \includegraphics[width=0.24\linewidth]{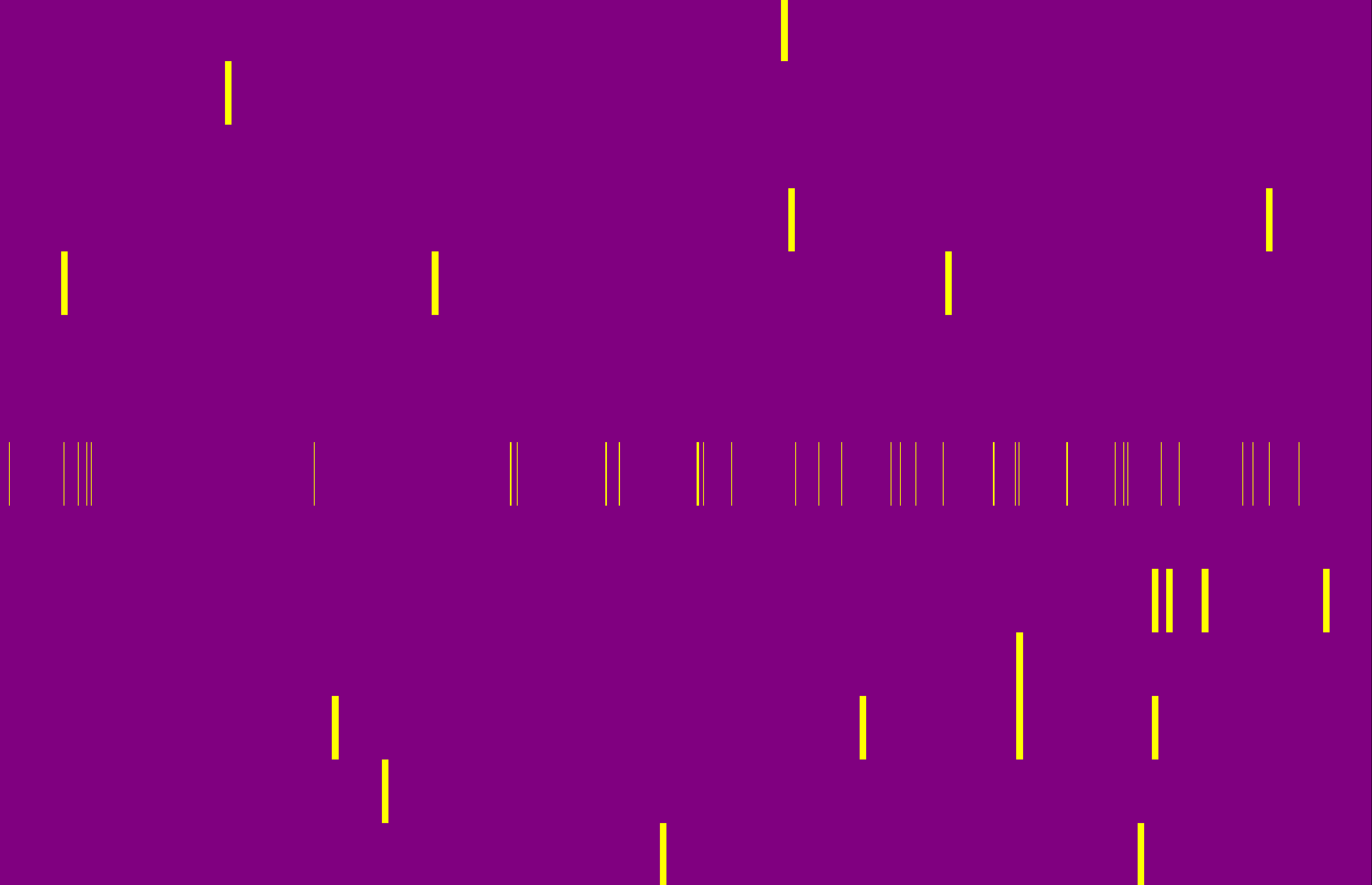}
    }
    \hfill
    \subfloat[]{%
        \includegraphics[width=0.24\linewidth]{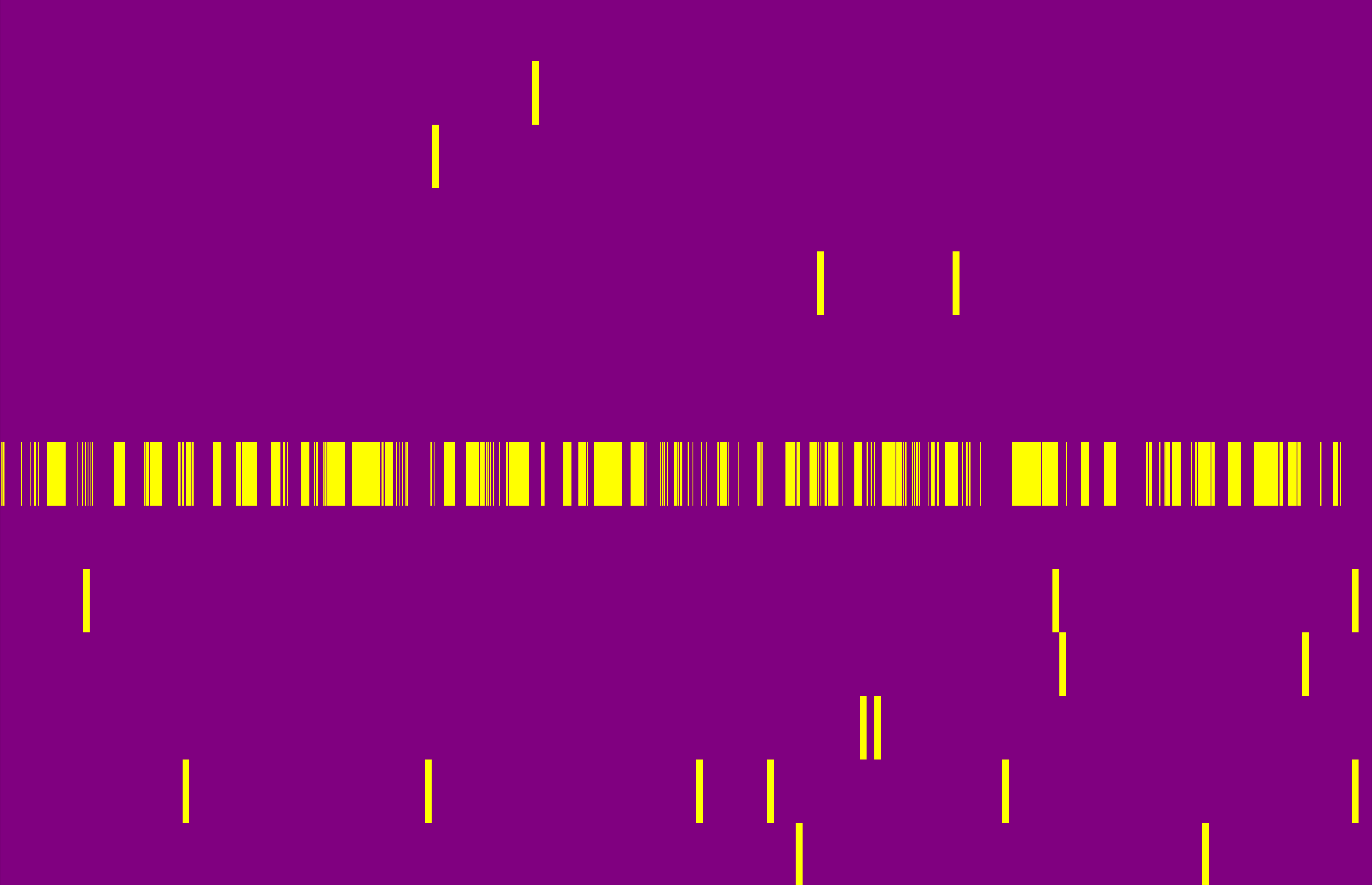}
      }\\[0pt]
    
    \subfloat[]{%
        \includegraphics[width=0.24\linewidth]{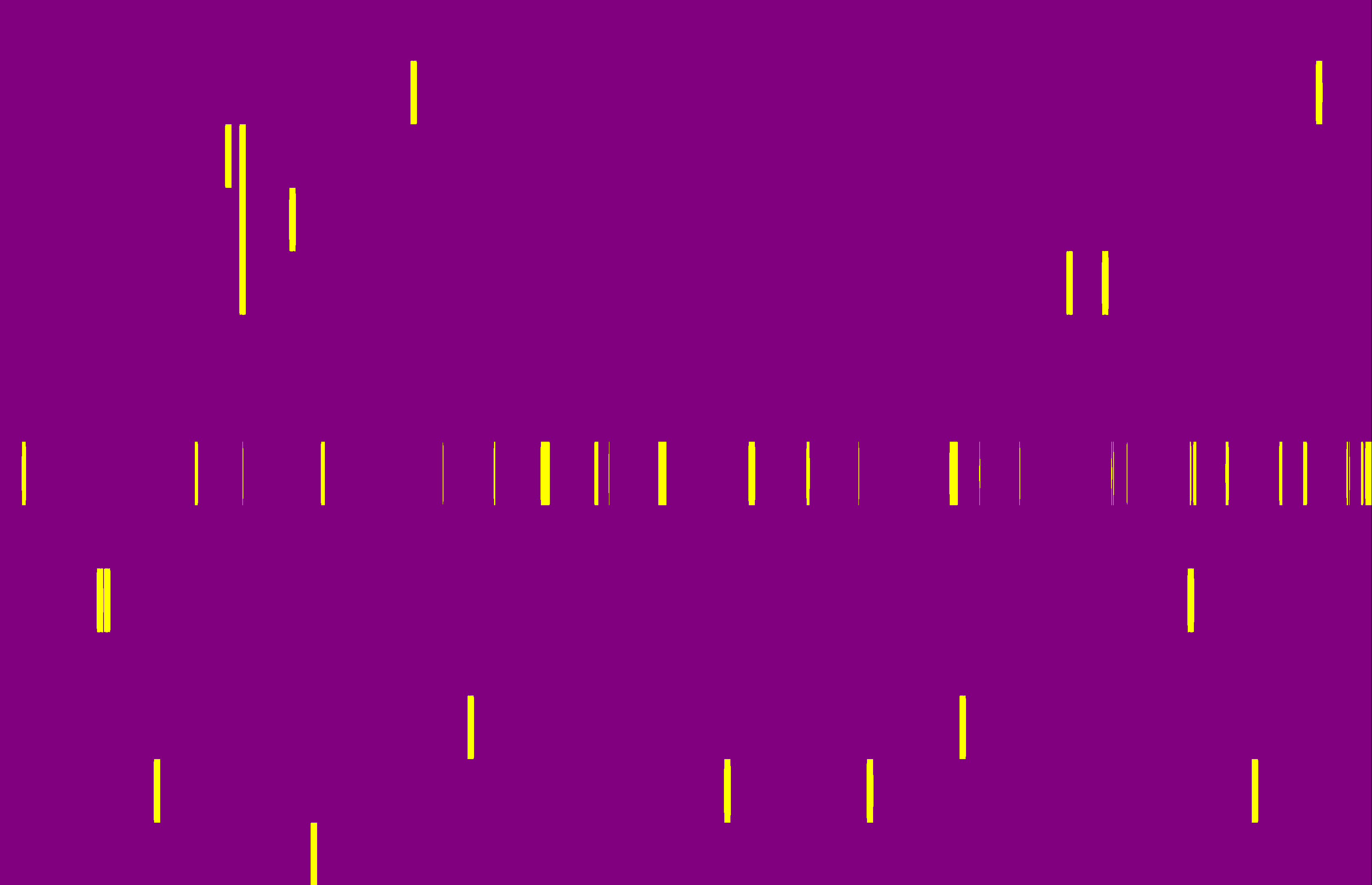}
      }
    \hfill
    \subfloat[]{%
        \includegraphics[width=0.24\linewidth]{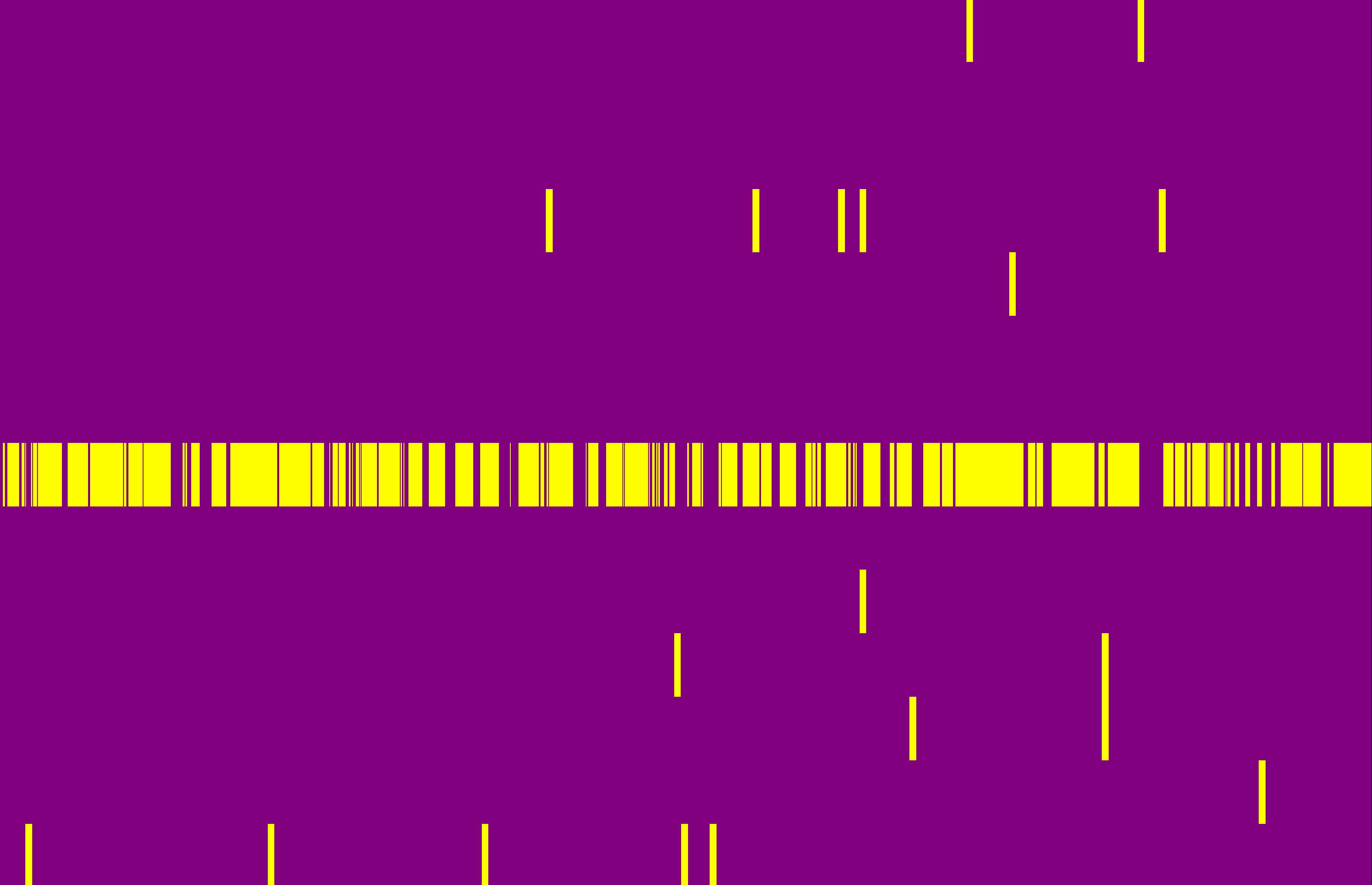}
  }
    \hfill
    \subfloat[]{%
        \includegraphics[width=0.24\linewidth]{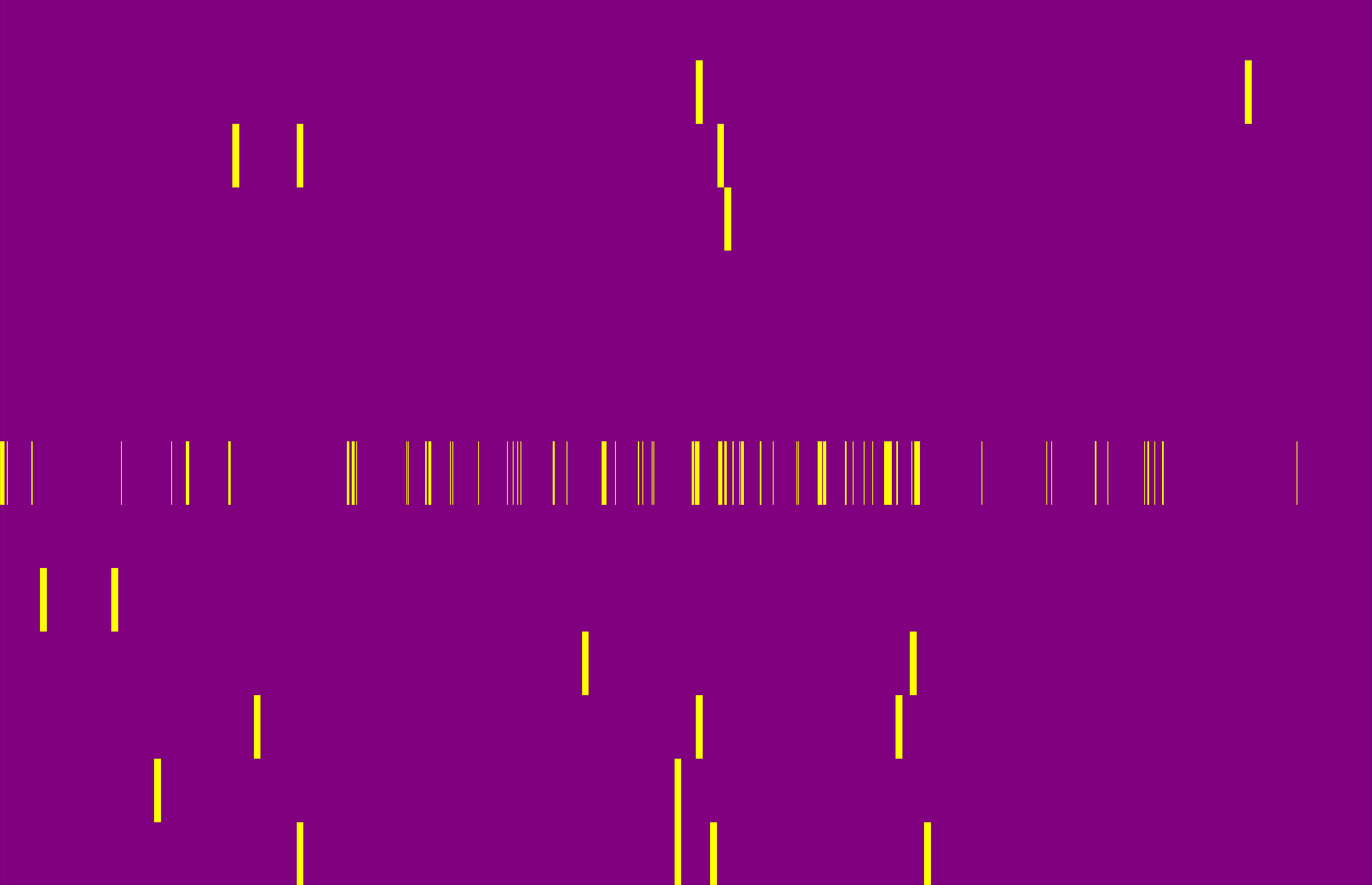}
      }
    \hfill
    \subfloat[]{%
        \includegraphics[width=0.24\linewidth]{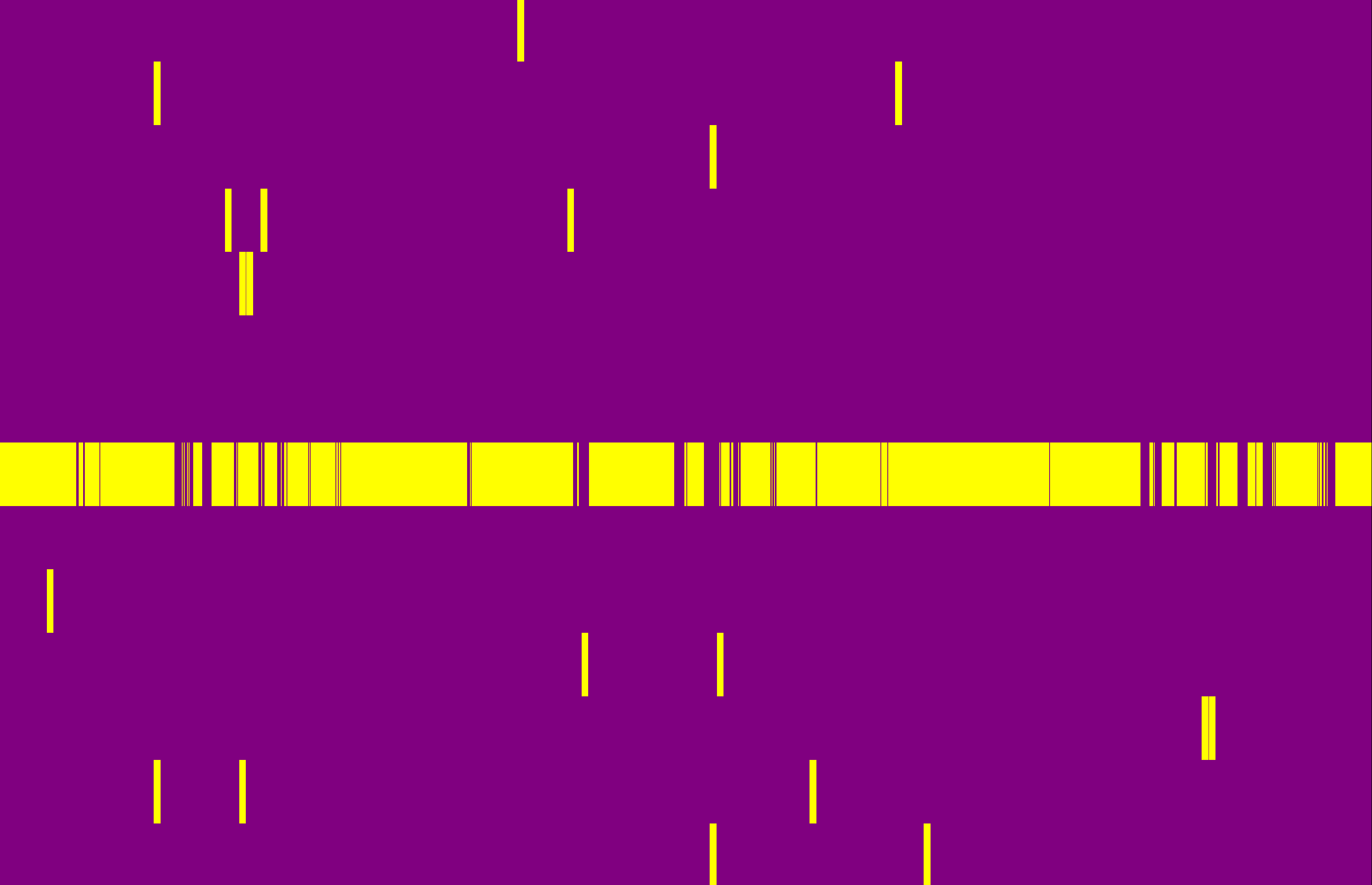}
      }\\[0pt]
    
    \subfloat[]{%
        \includegraphics[width=0.24\linewidth]{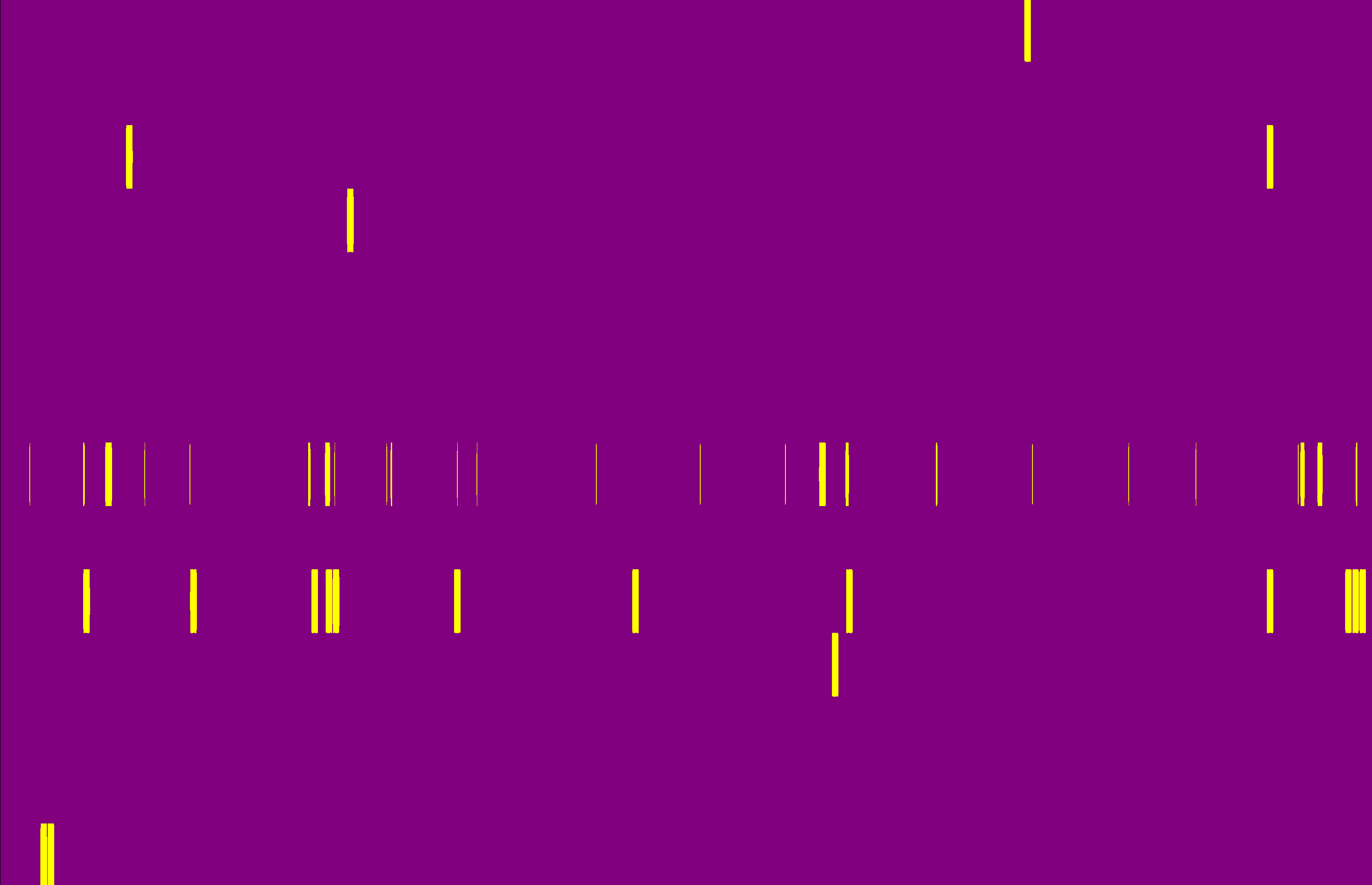}
      }
    \hfill
    \subfloat[]{%
        \includegraphics[width=0.24\linewidth]{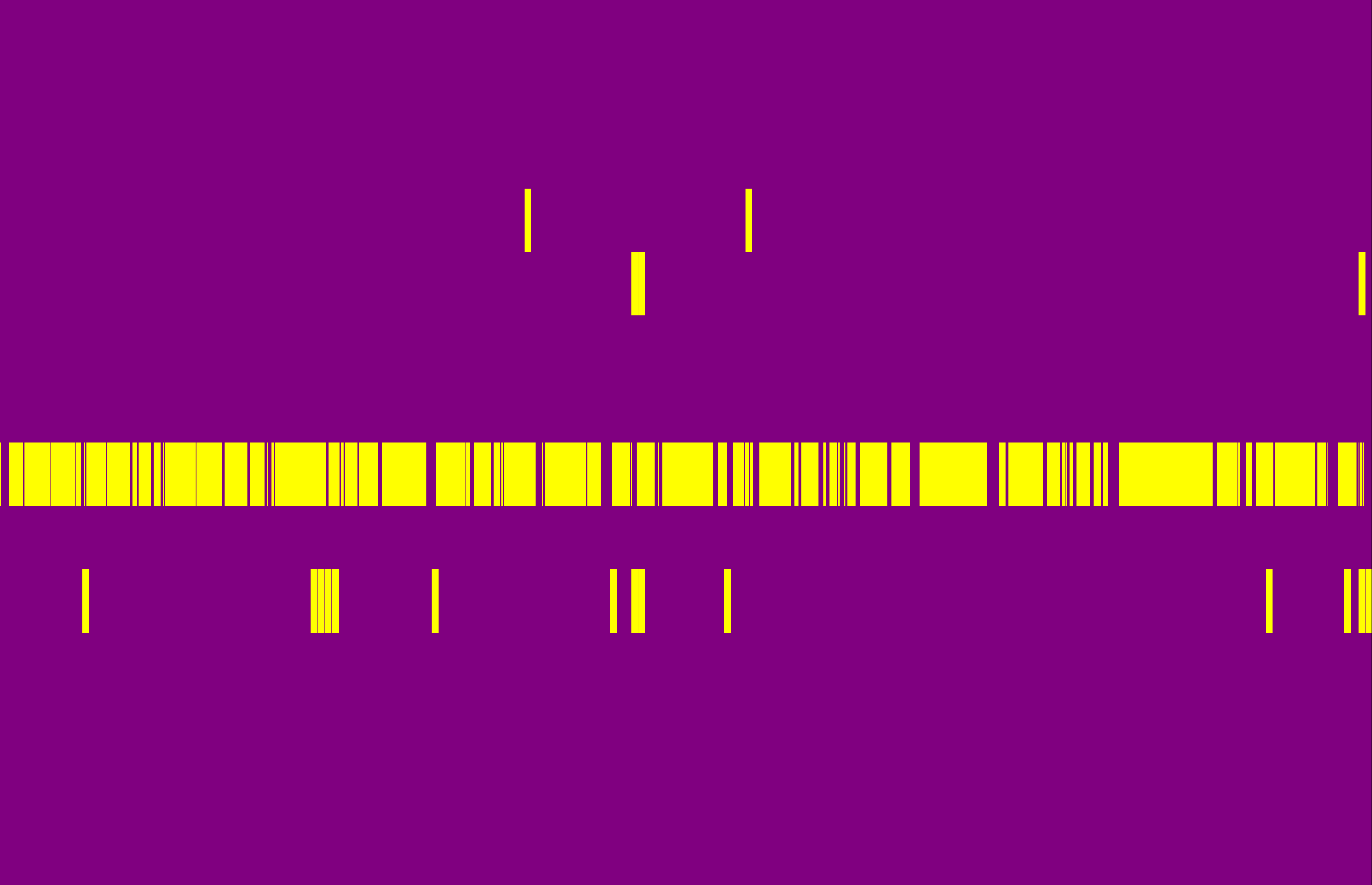}
     }
    \hfill
    \subfloat[]{%
        \includegraphics[width=0.24\linewidth]{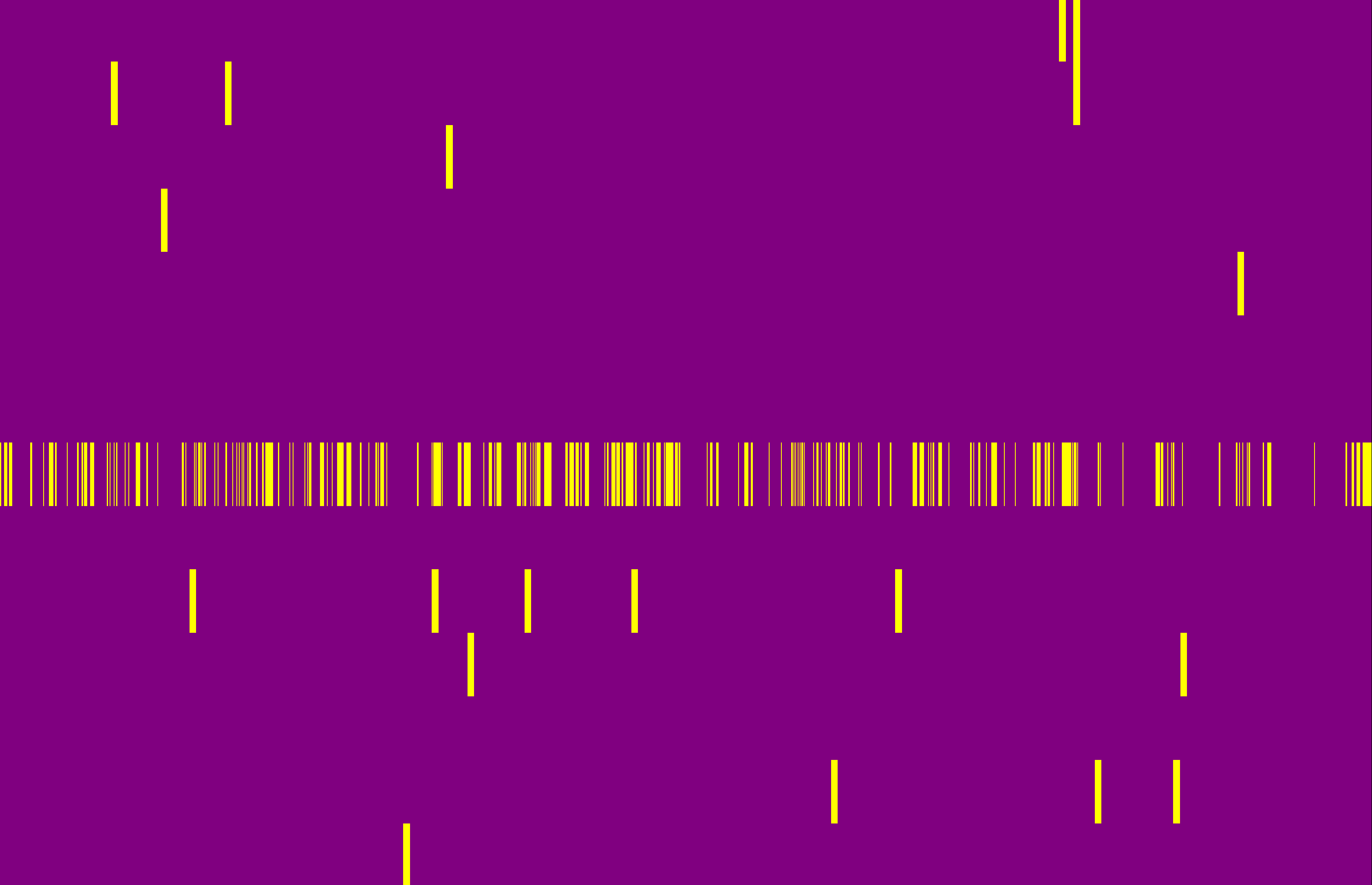}
       }
    \hfill
    \subfloat[]{%
        \includegraphics[width=0.24\linewidth]{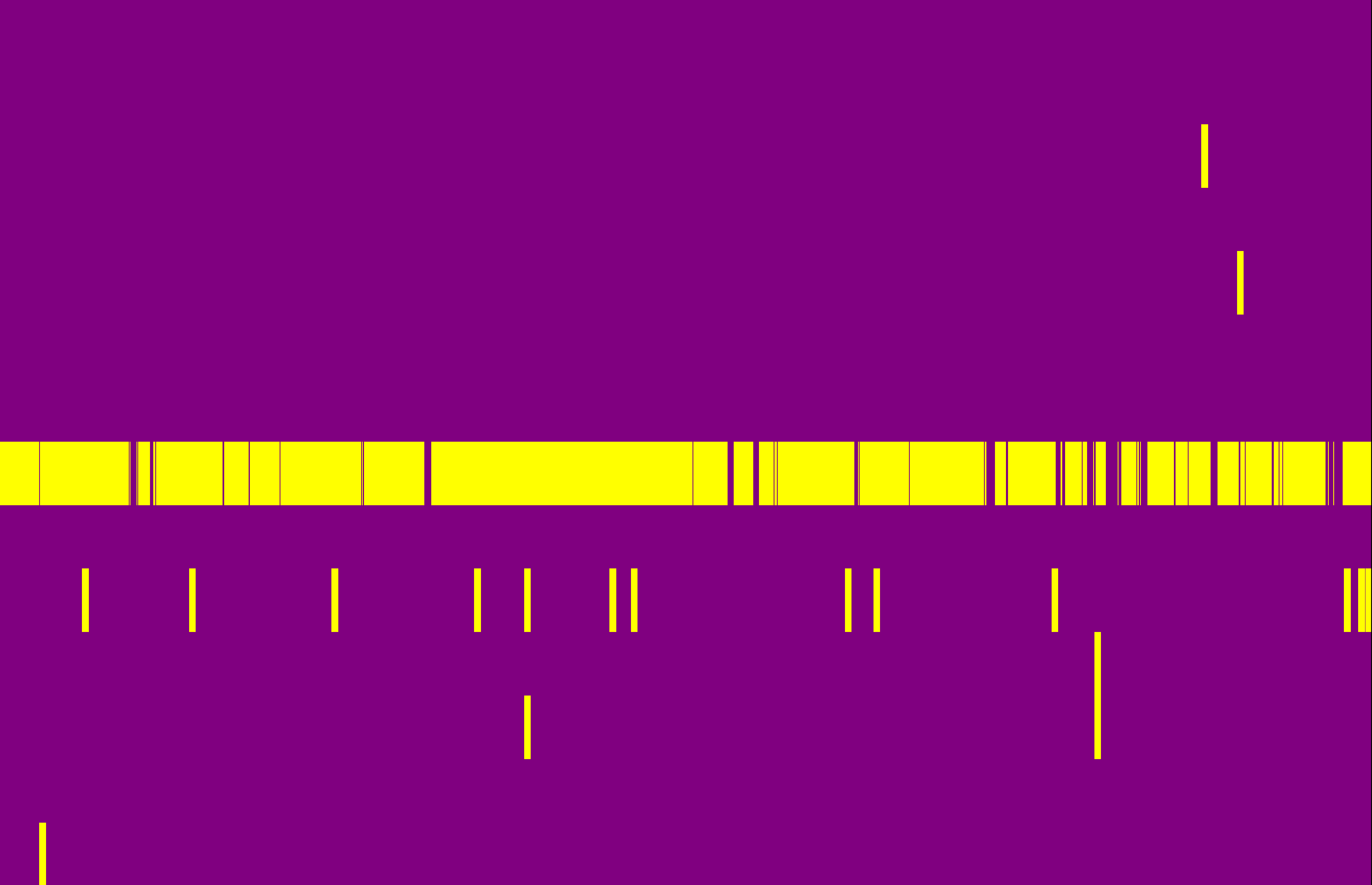}
       }
    
    \caption{Detection visualization results at $P_{fa}$=0.01. (a)\textasciitilde(d)Detection results of traditional supervised learning, unsupervised contrastive learning, supervised contrastive learning, and the proposed method on dataset \#280. (e)\textasciitilde(h)Detection results on dataset \#310. (i)\textasciitilde(l)Detection results on dataset \#311. (m)\textasciitilde(p) Detection results on dataset \#320.}
    \label{fig4}
\end{figure*}


\subsubsection{Generalization}

To demonstrate the generalization ability of the MDFG\_SCL method in complex scenarios, we combined the first six datasets from IPIX1993 for pretraining, enabling the model to learn radar echo representations across multiple scenes. The pretrained model was then evaluated on classification tasks using the remaining four datasets. Comparative experiments with traditional supervised learning, unsupervised contrastive learning, and supervised contrastive learning were conducted to validate the model's ability to capture underlying patterns in radar signal representations across different environments.

As shown in Fig. \ref{fig4}, MDFG\_SCL achieves the best detection performance at a $P_{fa}$ of 0.01, effectively identifying nearly all target samples. The next best performance comes from the unsupervised contrastive learning method RAVA-CL \cite{xia2023target}, which, despite poor results on single-dataset experiments, demonstrates strong generalization ability. This indicates that the pretraining process successfully captures implicit relationships between clutter and target samples, enabling effective representation of new data. In contrast, supervised learning and supervised contrastive learning perform poorly, highlighting that reliance solely on label constraints fails to capture all effective features within the same class. The proposed method, enhanced by shallow physical feature guidance during pretraining, benefits from knowledge-driven support, leading to superior performance on new data.


\section{Conclusion}

In this letter, we propose a method that leverages multi-domain shallow features to guide high-dimensional deep features, thereby improving maritime target detection. The proposed MDFG\_SCL effectively combines shallow and deep features to detect small targets in complex and variable scenarios. Experimental results demonstrate that the proposed detector, which integrates shallow and deep features, successfully addresses the generalization problems of existing detectors under changing environments, overcoming transfer challenges across different scenarios and achieving effective detection. In the future, we will continue to explore more beneficial shallow features based on physical models, including but not limited to temporal and spatial distribution information, to support high-dimensional features for more efficient and high-performance detection in intricate and dynamic environments.

\bibliographystyle{IEEEtran}
\bibliography{mybib}

\end{document}